\documentclass[conference]{IEEEtran}
\IEEEoverridecommandlockouts

\usepackage{cite}
\usepackage{amsmath,amssymb,amsfonts}
\usepackage{algorithmic}
\usepackage{graphicx}
\usepackage{textcomp}
\usepackage{xcolor}
\usepackage{tikz}
\usepackage{url}
\usetikzlibrary{positioning, arrows.meta, calc, fit, backgrounds, shadows.blur}
\usepackage{pgfplots}
\pgfplotsset{compat=1.18}
\definecolor{graftclr}{HTML}{166534}
\definecolor{baseclr}{HTML}{BFC4CC}
\usepackage{makecell}
\usepackage{booktabs}
\usepackage{multirow}
\usepackage{capt-of}
\definecolor{styleC}{RGB}{52,114,184}
\definecolor{hintC}{RGB}{220,128,52}
\definecolor{outC}{RGB}{120,120,120}
\definecolor{specC}{RGB}{86,156,106}
\newcommand{\red}[1]{{\color{red}#1}}
\def\BibTeX{{\rm B\kern-.05em{\sc i\kern-.025em b}\kern-.08em
    T\kern-.1667em\lower.7ex\hbox{E}\kern-.125emX}}
\begin{document}

\title{GRAFT: Grafted Reference Audio for Fine-grained
Pronunciation in Zero-shot Text-to-Speech}

\author{
\IEEEauthorblockN{Antonis Asonitis\textsuperscript{1,2,*}, Francesco Verdini\textsuperscript{1,3,*}, Aref Farhadipour\textsuperscript{1,4}, Vijeta Avijeet\textsuperscript{1},\\
Pierre-Edouard Honnet\textsuperscript{1}, Marzieh Razavi\textsuperscript{1}, Juan Pablo Zuluaga Gomez\textsuperscript{1}}
\IEEEauthorblockA{\textsuperscript{1}AGIGO \quad \textsuperscript{2}ETH Zurich \quad \textsuperscript{3}Sapienza University of Rome \quad \textsuperscript{4}University of Zurich}
\thanks{\textsuperscript{*}These authors contributed equally.}
}

\maketitle

\begin{abstract}
We present GRAFT, a per-word pronunciation conditioning mechanism for text-to-speech neural codec language modeling. Existing systems reach high intelligibility and naturalness but inherit the ambiguity of text and mispronounce rare proper nouns, loanwords and technical terms. Even phoneme-conditioned models offer no direct acoustic handle for per-word pronunciation. GRAFT controls the pronunciation of a chosen word from a short spoken sample of it, encoded with the model's own speech tokenizer and bound to the word's position in the prompt. Voice conversion during training-data construction disentangles the hint speaker from the target speaker, so the hint may come from any voice while the output stays in the target voice. In a blind English listening study, human raters rank GRAFT first by a clear margin, judging its rendering of the difficult word closest to a reference recording of that word. On a five-language objective benchmark, GRAFT reduces target-word phoneme error rate by 22–39\% over the identical text-only backbone and outperforms competitive open-source zero-shot systems, both phoneme- and text-conditioned, on target-word pronunciation, while preserving speaker similarity and naturalness.
\end{abstract}


\section{Introduction}
Early end-to-end text-to-speech systems mapped a phoneme sequence to a
waveform or a mel spectrogram \cite{tacotron2,fastspeech2}. The
explicit phonetic intermediate made the learning problem well
posed, since each phoneme corresponds to a small and largely
consistent set of acoustic realisations.

Recent systems extend a pretrained large
language model with a vocabulary of discrete speech tokens produced by
a neural audio codec, and continue training on paired text and speech
\cite{valle,llasa,qwen3tts,cosyvoice}. The strong text prior inherited
from the language model accelerates convergence and improves
intelligibility on common words. The downside is
that text is a highly compressed representation of what is spoken. The
orthographic form fixes the identity of the word but underdetermines
prosody, intonation, and the specific pronunciation in cases where
more than one is valid. For proper nouns, brand names, loanwords and
technical terms, the text prior often fails to faithfully align with the target acoustics.

\begin{figure}[t]
  \centering
  \includegraphics[width=\columnwidth]{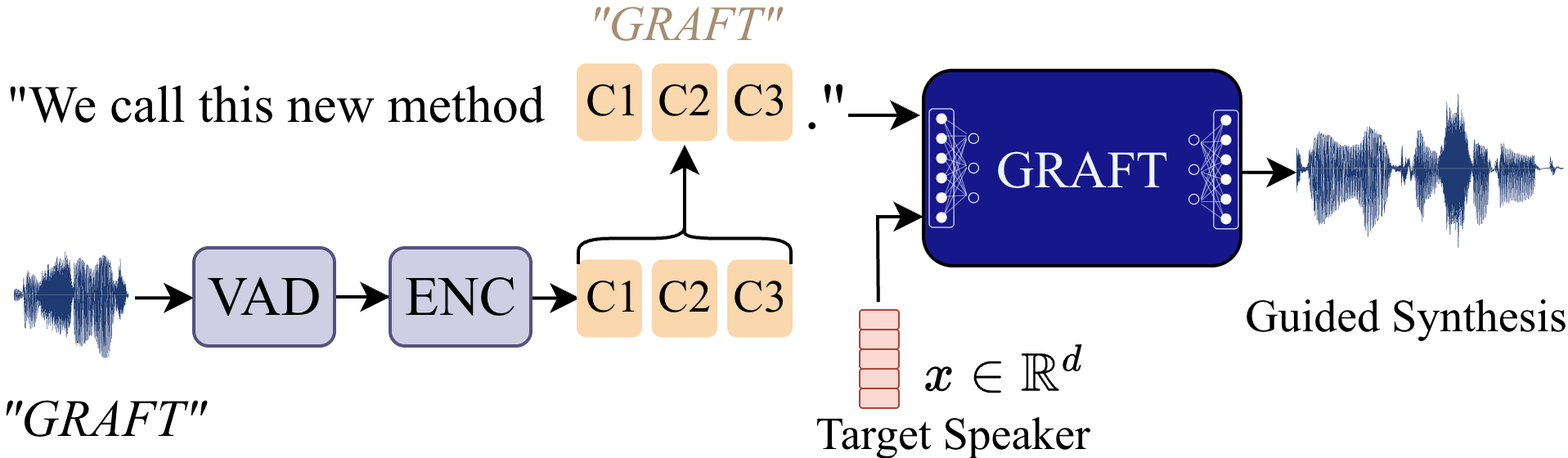}
  \caption{GRAFT at inference. A spoken example of the word
    (``GRAFT''), from any speaker, is silence-trimmed (VAD), encoded to
    codec tokens, and placed at that word's position in the text prompt.
    Conditioned on a separate target-speaker embedding, GRAFT
    synthesises the sentence in the target voice with the given
    pronunciation.}
  \label{fig:teaser}
\end{figure}

Models allowing jointly text and phonemes recover some control but do not fully resolve the problem. Most pronunciation control relies on rule-based grapheme-to-phoneme (G2P) lexicons~\cite{phonemizer,espeakng}, which assume a single canonical pronunciation per word and cannot capture words that admit several valid realisations. Such lexicons also generalise poorly to unseen words, the zero-shot case that matters most for rare names and loanwords. Neural phone recognisers such as ZIPA~\cite{zipa} and Wav2Vec2-Phoneme~\cite{wav2vec2phoneme} instead learn acoustic-to-phonetic mappings that generalise to new words and can convert an arbitrary recording into a phoneme sequence to prompt a phoneme-conditioned TTS. Even though this method offers tighter control, the fact that these phonemizers discard diacritical marks means the capacity to inject tonality or prosody is once again lost.

A more direct option is to condition on the least compressed representation that the architecture of a text-to-speech model already accepts, namely the model's own speech tokens. Controlling a word's pronunciation requires only a short recording of that word from any speaker. This needs no phonetic notation or lexicon entry, so a non-expert can prescribe a pronunciation simply by saying the word, and an acoustic example can also preserve the stress and tone that symbolic transcriptions discard. The TTS should then render the full target phrase in the target speaker voice while preserving the given pronunciation of the chosen word. We call this approach \emph{GRAFT} (Fig.~\ref{fig:teaser}).

Achieving this requires the model to disentangle the pronunciation in the reference recording from its speaker identity. The reference clip carries both, but only the pronunciation should transfer, while the target voice comes from a separate reference. This disentanglement follows from training on voice-conversion pairs, whose objective forces the model to carry the content across speakers and to ignore the speaker identity of the source. Once learned, the same disentanglement lets the model accept a per-word hint from any speaker and render the target phrase in a different target voice.


Our contributions are:
\begin{itemize}
  \item \textbf{GRAFT}, a per-word pronunciation control for text-to-speech neural codec language modeling that requires no additional parameters or architectural change, where a spoken example of a word replaces its text tokens and guides its pronunciation in any target voice across five different languages.\footnote{Samples available at \url{https://graft-tts.github.io/graft/}}
  \item Voice-converted training data is the key enabling factor, disentangling the hint's speaker from its pronunciation so a hint from any voice is rendered in the target voice.
  \item A multilingual difficult-word benchmark (five languages, openly licensed isolated references), with released checkpoints, code and evaluation tooling.
\end{itemize}

\section{Related Work}

\subsection{Neural codec language model TTS}
The dominant paradigm casts speech synthesis as next-token
prediction over discrete codec tokens, established by VALL-E
\cite{valle} and scaled by many successors
\cite{naturalspeech2,naturalspeech3,voicebox,cosyvoice,maskgct,llasa,qwen3tts}.
Most reuse a pretrained text-only LLM and extend its vocabulary with
codec indices \cite{llasa,qwen3tts}, which is the source of both the
strong text prior and the pronunciation failure modes that GRAFT
targets. They clone a voice zero-shot by conditioning on a short acoustic
example \cite{speartts}. GRAFT inherits this architecture unchanged and
adds a per-word audio conditioning slot consumed at inference, with no
new parameters: the hint is encoded by the model's existing codec
tokenizer and spliced into the prompt, leaving inference cost unchanged
apart from the few extra tokens. Closest to our mechanism,
WESCON~\cite{wescon} splices a reference acoustic prompt at a chosen
word to control its emotional expression. GRAFT applies similar
per-word conditioning for pronunciation and adds voice-conversion
training so the example may come from a different speaker than the
target.

\subsection{Phoneme-conditioned TTS}
Phoneme conditioning is the classical interface for pronunciation
control, from sequence-to-sequence models
\cite{tacotron2,fastspeech2,glowtts,vits} to neural codec language models
that mix phonemes into the prompt \cite{naturalspeech3,voicebox}. All
commit to a fixed phonemic vocabulary \cite{phonemizer,espeakng}, which
restricts coverage. Improved text-side front-ends \cite{pngbert,charsiug2p}
predict pronunciations more accurately but still emit symbols from a
fixed inventory. Per-word fixes read an external
lexicon entry~\cite{neurallexicon} or edit the model's weights for the
target word~\cite{sonoedit}, but both require a written phonetic
specification. GRAFT removes this commitment by conditioning on an
instance of the desired pronunciation rather than a symbolic
transcription.

\subsection{Forced alignment}
Forced alignment locates word boundaries from a transcript, via HMM/GMM models such as the Montreal Forced Aligner~\cite{mfa} or CTC-based neural aligners~\cite{ctcseg}. A second family avoids a dedicated alignment model and instead recovers timing from the cross-attention of an existing ASR or TTS decoder: the attention weights that link each output token to the audio frames are collapsed into a monotonic path with dynamic time warping, as in Whisper's native word timestamps and tools such as whisper-timestamped~\cite{whispertimestamped} and stable-ts~\cite{stablets}, a hybrid variant instead transfers timing from a separate CTC model onto the transcript~\cite{whisperx} GRAFT uses the Qwen3-TTS aligner~\cite{qwen3tts} only in data preparation, to cut the per-word reference clips. None is needed at inference, where the input is already isolated.

\subsection{Voice conversion}
Voice conversion renders an utterance in a target speaker's voice
while preserving content, with recent zero-shot systems including
kNN-VC \cite{knnvc}, diffusion approaches \cite{diffvc} and
Seed-VC \cite{seedvc}, which we use in data preparation. Its
speaker-content factorisation has been studied via
information-bottleneck~\cite{autovc}, self-supervised content
units~\cite{contentvec} and perturbation training~\cite{nansy}. Prior
work treats voice conversion as the goal. GRAFT uses it only for training data construction, in order to convert each hint to a different speaker, which teaches
GRAFT to copy a word's pronunciation and prosody while ignoring the hint speaker, so
at inference the reference may come from any voice. In this work, we show that this disentanglement is necessary to achieve natural-sounding speech.

\section{Method}
\label{sec:method}

\subsection{Preliminaries}
The base system is a neural codec language model that autoregressively
predicts $N$-codebook codec tokens from a serialised text prompt. We
write a token sequence as $\mathbf{c} \in \mathbb{Z}^{T \times N}$ ($T$
frames at $f$~Hz). We use \textbf{Qwen3-TTS-12Hz-0.6B-Base}
\cite{qwen3tts} throughout ($N{=}16$, $f{=}12.5$~Hz). It exposes a codec
encoder $\mathrm{Enc}(\cdot)$ ($\mathbf{c} = \mathrm{Enc}(\mathbf{a})$ on
a 24~kHz waveform), a decoder $\mathrm{Dec}(\cdot)$, a text tokenizer, and
special tokens. GRAFT uses two slot delimiters,
$\langle Q\rangle/\langle/Q\rangle$ (style references) and
$\langle B\rangle/\langle/B\rangle$ (per-word grafts), with a placeholder
$\square$.

\begin{table}[!t]
\centering
\caption{Training mix by language. Counts are pooled over the corpora feeding
each language and use k ($10^3$) and M ($10^6$). Corpora: VCTK~\cite{vctk}~(V),
LibriTTS~\cite{libritts}~(L), MLS~\cite{mls}~(M) and Common
Voice~\cite{commonvoice}~(C). ``\#Spk'' is unique speaker IDs, ``\#Utt''
utterances after preprocessing.}
\label{tab:training-data}
\small
\setlength{\tabcolsep}{5pt}
\renewcommand{\arraystretch}{1.15}
\begin{tabular*}{\columnwidth}{@{\extracolsep{\fill}}llccc@{}}
\toprule
\textbf{Lang.} & \textbf{Corpora} & \textbf{\#Spk} & \textbf{Hours} & \textbf{\#Utt} \\
\midrule
EN    & V, L, C & $12.9$k & $1.5$k & $1.1$M \\
DE    & M, C    & $0.8$k   & $2.0$k & $0.9$M \\
FR    & M, C    & $9.0$k  & $1.9$k & $0.8$M \\
ES    & M, C    & $4.9$k  & $1.4$k & $0.6$M \\
IT    & M, C    & $1.5$k  & $0.5$k & $0.2$M \\
\midrule
\textbf{Total} & --- & $\mathbf{28.4}$k & $\mathbf{7.3}$k & $\mathbf{3.2}$M \\
\bottomrule
    \end{tabular*}
\end{table}

\subsection{Per-word audio binding}
GRAFT conditions on a tuple
$\bigl(t,\, \mathbf{r},\, \{(w_k, \mathbf{a}_k)\}_{k=1}^{K}\bigr)$
where $t$ is the target text, $\mathbf{r}$ is a target-voice
reference clip and each $(w_k, \mathbf{a}_k)$ is a target word paired
with an isolated audio recording of that word. The reference clip
and the per-word clips are encoded by the base model's own codec,
$\mathbf{c}^{\mathrm{ref}} = \mathrm{Enc}(\mathbf{r})$ of length
$T_r$, and $\mathbf{c}^{(k)} = \mathrm{Enc}(\mathbf{a}_k)$ of length
$T_k$.

Let the target text be $t = m_1 \cdots m_M$ and
$\mathcal{G} = \{i : m_i = w_k \text{ for some } k\}$ index the grafted
words. The prompt concatenates a fixed role prefix, the style slot
$[\langle Q\rangle, \square^{T_r}, \langle/Q\rangle]$, and a body in
which each ungrafted word ($i \notin \mathcal{G}$) keeps its text tokens
while each grafted word ($i \in \mathcal{G}$) contributes the graft slot
$[\langle B\rangle, \square^{T_k}, \langle/B\rangle]$ in place.
A grafted word is thus represented by its codec tokens \emph{instead
of} its text, never in addition to it.
The placeholder text embeddings are replaced at the corresponding
positions by the codec embeddings
$\mathbf{e}^{\mathrm{codec}}_{t} = \sum_{n=0}^{N-1} E_n\!\bigl(\mathbf{c}[t, n]\bigr)$,
summed across all $N{=}16$ residual codebooks, where $E_n$ is the
embedding table for the $n$-th codebook. The talker transformer then
generates the remaining codec frames autoregressively, conditioned on
this composite input.

\subsection{Training data construction}
\label{sec:data}

\begin{figure}[!t]
  \centering
    \includegraphics[width=0.8\columnwidth]{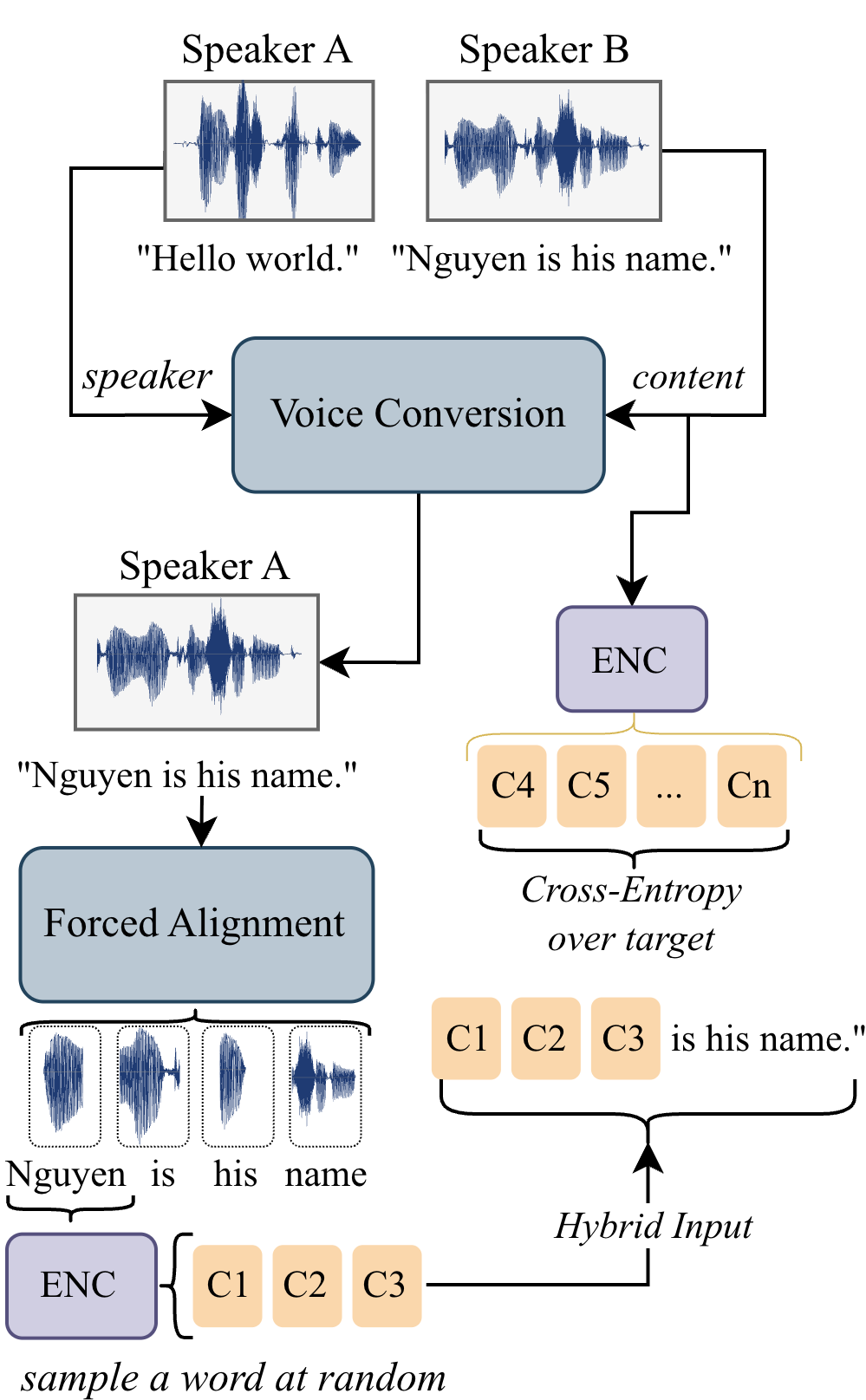}
      \caption{Training-data construction. From a Speaker~A utterance we make
  a voice-converted copy in a different voice, force-align it to find word
  boundaries, and re-encode the target word (``Nguyen'') with the codec as
  a per-word graft. The graft replaces that word's text in
  the prompt, and the model is trained to predict the codec of the
  \emph{original} Speaker~A utterance. Since the graft is in a different
  voice from the target, the model learns to copy pronunciation while
  ignoring the hint speaker.}
  \label{fig:training}
\end{figure}

We construct training samples that match the inference-time use of
GRAFT, using only the public read-speech and crowd-sourced corpora of
Table~\ref{tab:training-data} and no additional human-supplied audio.

For every training utterance $u$ with original waveform $\mathbf{a}^u$ and
transcript $t^u$, we produce a parallel voice-converted version
$\hat{\mathbf{a}}^u = \mathrm{SeedVC}_{D}(\mathbf{a}^u,\mathbf{r}^u)$,
where $\mathbf{r}^u$ is a reference clip from a random other speaker and
Seed-VC \cite{seedvc} runs with $D = 20$ steps, so the utterance is said
in a different voice. We force-align
$\hat{\mathbf{a}}^u$ to $t^u$, cut each word $i$, pad it with $0.3$~s
of silence on each side, encode it with the base codec, and discard
the codec frames in the padding. The result is a per-word codec
sequence $\mathbf{c}^{u,i}\!\in\!\mathbb{Z}^{T^{u,i}\times N}$ for
every word, each in a speaker different from the original.

The training target is always the original waveform's codec,
$\mathbf{c}^{u,\mathrm{tgt}} = \mathrm{Enc}(\mathbf{a}^u)$. The
voice-converted utterance is used as input for per-word reference clips spliced into the prompt and  never as target. Therefore, GRAFT's output quality is not limited by the voice conversion model.

\begin{table}[!t]
\caption{Ablation of training-data choices, evaluated on a separate set
of $100$ difficult English words manually selected from the ACL~60/60
corpus~\cite{acl6060}. \emph{no~VC} removes voice conversion from
training-data construction and \emph{no aug.}\ removes the per-word hint
augmentations. SSIM is speaker similarity to the target voice. Best per
column in bold. $^{*}$no-VC reaches the lowest D-PER only by copying the
hint tokens straight through, which collapses speaker similarity, so it
is a degenerate solution rather than the strongest system.}
\label{tab:ablation}
\centering
\small
\setlength{\tabcolsep}{6pt}
\renewcommand{\arraystretch}{1.1}
\begin{tabular}{@{}lcccc@{}}
\toprule
\textbf{System} & \textbf{D-PER}\,$\downarrow$ & \textbf{UTMOS}\,$\uparrow$
 & \textbf{WER}\,$\downarrow$ & \textbf{SSIM}\,$\uparrow$ \\
\midrule
GRAFT (no aug.) & $0.25$ & $4.21$ & $0.08$ & $0.67$ \\
GRAFT (no VC) & $0.16^{*}$ & $3.89$ & $0.18$ & $0.20$ \\
\textbf{GRAFT} & $0.19$ & $\mathbf{4.43}$ & $\mathbf{0.05}$ & $\mathbf{0.68}$ \\
\bottomrule
\end{tabular}
\end{table}

\begin{table*}[!t]
\caption{Difficult-word benchmark ($500$ words per language, one run per phrase). Phoneme baselines (StyleTTS2, MaskGCT, Matcha-TTS, Piper) receive the hint \emph{audio} transcribed to espeak phonemes (wav2vec2-espeak~\cite{wav2vec2phoneme}), the same hint pronunciation GRAFT gets but in symbolic form. Text-input systems (XTTS-v2, F5-TTS, CosyVoice2) get only the spelled carrier, and GRAFT gets the hint as audio. Systems appear only for languages they support. D-PER is mean${\pm}$std over words, and D-PER$_{10\%}$ averages the hardest $10\%$. $\downarrow$/$\uparrow$ lower/higher is better. $^{\dagger}$single-speaker (excluded from the SSIM ranking). Best \textbf{bold}, second \underline{underlined}.
The English Bradley-Terry study is in Fig.~\ref{fig:elo}.}
\label{tab:difficult}
\centering
\footnotesize
\setlength{\tabcolsep}{4pt}
\renewcommand{\arraystretch}{0.95}
\begin{tabular*}{\textwidth}{@{\extracolsep{\fill}}c l cccccc@{}}
\toprule
 & & \multicolumn{3}{c}{\textbf{Pronunciation}} & \multicolumn{3}{c}{\textbf{General}} \\
\cmidrule(lr){3-5}\cmidrule(lr){6-8}
 & \textbf{System} & \textbf{D-PER}\,$\downarrow$ & \textbf{D-PER$_{10\%}$}\,$\downarrow$ & \textbf{WS}\,$\uparrow$ & \textbf{WER}\,$\downarrow$ & \textbf{UTMOS}\,$\uparrow$ & \textbf{SSIM}\,$\uparrow$ \\
\midrule
\multirow{9}{*}{\rotatebox{90}{\textbf{English}}} & Qwen3-TTS-0.6B &$\underline{0.37}\,{\scriptstyle\pm0.42}$ & $1.16$ & $\underline{0.91}$& $\mathbf{0.02}$ & $\underline{4.46}$ & $0.73$  \\
 & StyleTTS2 (ph.) & $0.65\,{\scriptstyle\pm0.28}$ & $1.11$ & $0.83$& $0.17$ & $4.35$ & $0.62$  \\
 & MaskGCT (ph.) & $0.46\,{\scriptstyle\pm0.34}$ & $1.15$ & $0.88$& $0.17$ & $4.22$ & $\mathbf{0.80}$  \\
 & Matcha (ph.) & $0.43\,{\scriptstyle\pm0.34}$ & $1.11$ & $0.89$& $0.06$ & $4.34$ & $0.15^{\dagger}$  \\
 & Piper (ph.) & $0.48\,{\scriptstyle\pm0.36}$ & $1.12$ & $0.86$& $0.07$ & $4.34$ & $0.12^{\dagger}$  \\
 & XTTS-v2 & $0.41\,{\scriptstyle\pm0.38}$ & $1.20$ & $0.90$& $\mathbf{0.02}$ & $4.19$ & $0.67$  \\
 & F5-TTS & $0.38\,{\scriptstyle\pm0.40}$ & $\underline{1.10}$ & $0.89$& $\mathbf{0.02}$ & $4.35$ & $\underline{0.79}$  \\
 & CosyVoice2 & $0.39\,{\scriptstyle\pm0.39}$ & $1.16$ & $0.90$& $\underline{0.03}$ & $\underline{4.46}$ & $0.71$  \\
 & \textbf{GRAFT} & $\mathbf{0.29}\,{\scriptstyle\pm0.16}$ & $\mathbf{0.81}$ & $\mathbf{0.92}$& $0.09$ & $\mathbf{4.47}$ & $0.76$  \\
\midrule
\multirow{5}{*}{\rotatebox{90}{\textbf{German}}} & Qwen3-TTS-0.6B &$\underline{0.49}\,{\scriptstyle\pm0.33}$ & $\underline{1.13}$ & $\underline{0.90}$& $\mathbf{0.07}$ & $\mathbf{4.03}$ & $\underline{0.69}$  \\
 & MaskGCT (ph.) & $0.56\,{\scriptstyle\pm0.38}$ & $1.28$ & $0.87$& $0.17$ & $3.07$ & $\mathbf{0.79}$  \\
 & Piper (ph.) & $0.55\,{\scriptstyle\pm0.33}$ & $1.19$ & $0.87$& $0.13$ & $3.36$ & $0.14^{\dagger}$  \\
 & XTTS-v2 & $0.50\,{\scriptstyle\pm0.39}$ & $1.25$ & $\underline{0.90}$& $\underline{0.11}$ & $3.18$ & $\underline{0.69}$  \\
 & \textbf{GRAFT} & $\mathbf{0.30}\,{\scriptstyle\pm0.23}$ & $\mathbf{0.76}$ & $\mathbf{0.94}$& $\underline{0.11}$ & $\underline{3.99}$ & $0.68$  \\
\midrule
\multirow{5}{*}{\rotatebox{90}{\textbf{French}}} & Qwen3-TTS-0.6B &$\underline{0.41}\,{\scriptstyle\pm0.37}$ & $1.20$ & $\underline{0.89}$& $\mathbf{0.05}$ & $\underline{3.64}$ & $0.70$  \\
 & MaskGCT (ph.) & $0.56\,{\scriptstyle\pm0.35}$ & $1.25$ & $0.86$& $0.18$ & $2.80$ & $\mathbf{0.81}$  \\
 & Piper (ph.) & $0.55\,{\scriptstyle\pm0.34}$ & $1.23$ & $0.85$& $0.14$ & $3.40$ & $0.11^{\dagger}$  \\
 & XTTS-v2 & $0.43\,{\scriptstyle\pm0.34}$ & $\underline{1.11}$ & $0.88$& $\underline{0.06}$ & $3.04$ & $0.70$  \\
 & \textbf{GRAFT} & $\mathbf{0.28}\,{\scriptstyle\pm0.18}$ & $\mathbf{0.77}$ & $\mathbf{0.92}$& $0.07$ & $\mathbf{3.72}$ & $\underline{0.71}$  \\
\midrule
\multirow{4}{*}{\rotatebox{90}{\textbf{Spanish}}} & Qwen3-TTS-0.6B &$\underline{0.38}\,{\scriptstyle\pm0.32}$ & $1.02$ & $\underline{0.90}$& $\mathbf{0.02}$ & $\underline{3.76}$ & $\mathbf{0.69}$  \\
 & Piper (ph.) & $0.56\,{\scriptstyle\pm0.33}$ & $1.16$ & $0.85$& $0.10$ & $2.53$ & $0.08^{\dagger}$  \\
 & XTTS-v2 & $0.40\,{\scriptstyle\pm0.29}$ & $\underline{0.99}$ & $0.89$& $\underline{0.04}$ & $3.08$ & $0.65$  \\
 & \textbf{GRAFT} & $\mathbf{0.29}\,{\scriptstyle\pm0.20}$ & $\mathbf{0.68}$ & $\mathbf{0.91}$& $0.09$ & $\mathbf{3.80}$ & $\underline{0.66}$  \\
\midrule
\multirow{4}{*}{\rotatebox{90}{\textbf{Italian}}} & Qwen3-TTS-0.6B &$0.40\,{\scriptstyle\pm0.28}$ & $0.97$ & $\underline{0.89}$& $\mathbf{0.06}$ & $\underline{3.68}$ & $\underline{0.70}$  \\
 & Piper (ph.) & $0.58\,{\scriptstyle\pm0.27}$ & $1.13$ & $0.81$& $0.15$ & $3.40$ & $0.13^{\dagger}$  \\
 & XTTS-v2 & $\underline{0.38}\,{\scriptstyle\pm0.27}$ & $\underline{0.94}$ & $0.88$& $\underline{0.09}$ & $2.97$ & $\mathbf{0.71}$  \\
 & \textbf{GRAFT} & $\mathbf{0.25}\,{\scriptstyle\pm0.17}$ & $\mathbf{0.66}$ & $\mathbf{0.91}$& $0.11$ & $\mathbf{3.73}$ & $\mathbf{0.71}$  \\
\bottomrule
\end{tabular*}
\end{table*}


To assemble a training input, we walk over the words of $t^u$ and
decide for each word whether to insert a per-word graft. A word is
eligible if its aligned duration is at least $0.04$~s and its
surface form is at least four characters long. Each eligible word
$i$ is selected independently with probability
$p_{\mathrm{hint}} = 0.5$. For every selected word, the actual
graft codec sequence is sampled according to
\begin{equation}
\mathbf{c}^{(i)} \;=\;
\begin{cases}
\mathbf{c}^{u,i}                          & \text{w.p. } (1{-}p_s)(1{-}p_a),\\
\mathrm{Aug}\bigl(\mathbf{c}^{u,i}\bigr)  & \text{w.p. } (1{-}p_s)\,p_a,\\
\mathbf{c}^{u',i}                          & \text{w.p. } p_s(1{-}p_a),\\
\mathrm{Aug}\bigl(\mathbf{c}^{u',i}\bigr)  & \text{w.p. } p_s\,p_a,
\end{cases}
\label{eq:hint-sampling}
\end{equation}
with $p_s = 0.3$ and $p_a = 0.5$. The first two cases keep the
voice-converted clip from the current utterance, with or without
augmentation. The last two swap in the same word from a different
utterance $u'$. The swap matters because an in-context graft is cut
from the very sentence being predicted and is therefore coarticulated
and prosodically consistent with the carrier, whereas at inference the
reference is an isolated recording. Sourcing the graft from a different
utterance breaks this coarticulation and matches the isolated-word
prompts seen at inference.

Augmented variants are precomputed offline: each per-word clip is
decoded, perturbed, and re-encoded, so
$\mathrm{Aug}(\mathbf{c})=\mathrm{Enc}(\mathrm{perturb}(\mathrm{Dec}(\mathbf{c})))$
in Eq.~\eqref{eq:hint-sampling}. The perturbations cover additive
noise, level and loudness jitter, pitch and speed changes,
telephone-band filtering, compression, reverberation and variable
lead/trail silence, alone and in combinations emulating phone,
quiet-room, far-microphone and loud-close conditions. They broaden the
conditioning distribution along the two axes that matter at inference:
imperfect word boundaries, where the variable lead/trail silence makes
the model robust to spans that are cut loosely, and consumer-microphone
recording in noisy rooms.

\subsection{Training objective}
We fine-tune the base model with the standard next-token cross-entropy loss on the talker codec tokens plus an auxiliary sub-talker loss weighted $\lambda_{\mathrm{sub}} = 0.3$, $\mathcal{L} = \mathcal{L}_{\mathrm{talker}} + \lambda_{\mathrm{sub}}\,\mathcal{L}_{\mathrm{sub-talker}}$.
Training runs over the full mix of Table~\ref{tab:training-data} with
effective batch size $192$, AdamW \cite{adamw}, learning rate
$5{\times}10^{-6}$, weight decay $0.01$, a cosine schedule and $0.02$
warmup, taking a total of 48 hours on a single 96~GB GPU. The ablations
use the identical schedule and data count, isolating the
training-data choice.

\subsection{Inference}
At inference the inputs are only $t$, $\mathbf{r}$ and the
$(w_k, \mathbf{a}_k)$ pairs, with no forced alignment since each
$\mathbf{a}_k$ is already isolated. We use standard autoregressive
sampling, and trimming each clip with a voice activity
detector (VAD)~\cite{silerovad} before encoding brings it closer to the
training distribution.

\section{Experimental Setup}
\label{sec:setup}

\subsection{Difficult-word benchmark}
We evaluate on a new multilingual benchmark of difficult words. For each
of five languages (English, German, French, Spanish, Italian) we select
the rarest words by \texttt{wordfreq} Zipf score (band $1.3$--$3.3$) from
the openly licensed Wikimedia Lingua Libre collection of isolated
single-word recordings ($500$ per language, $2{,}500$ total), place each
in a natural carrier phrase, and use its clean isolated human recording as
the per-word hint. Because the recordings are isolated, no forced alignment
or cropping is needed, and every item is openly licensed
(CC0/CC-BY/CC-BY-SA), so the benchmark is redistributed with its audio. Each
phrase is synthesised once per system. We report target-word D-PER
(mean${\pm}$std) with its mean over the hardest $10\%$, alongside WER,
UTMOS and speaker similarity.



\subsection{Compared systems}
The text-only base, Qwen3-TTS-12Hz-0.6B-Base \cite{qwen3tts},
is the principal baseline, receiving only the target text and the style
reference with no per-word audio interface. We additionally compare
against seven open-source zero-shot systems, grouped by conditioning
interface. Four take a phonemic front-end:
StyleTTS2~\cite{styletts2} (diffusion),
MaskGCT~\cite{maskgct} (masked codec LM),
Matcha-TTS~\cite{matcha} (flow matching) and
Piper~\cite{piper} (VITS). Three are text-input:
XTTS~v2~\cite{xtts},
F5-TTS~\cite{f5tts} and
CosyVoice2~\cite{cosyvoice2}.
All receive the same carrier and speaker reference as GRAFT, so the only
difference is the model and its interface. The phoneme-input systems do
not merely get the spelled word: we transcribe the hint \emph{audio} to
espeak phonemes (wav2vec2-espeak~\cite{wav2vec2phoneme}) and substitute
that for the target word, the matched symbolic counterpart to GRAFT's
audio hint. The text-input systems cannot consume phonemes and receive
only the spelled carrier. The methods designed specifically for per-word
correction, Neural Lexicon Reader~\cite{neurallexicon} and
SonoEdit~\cite{sonoedit}, both condition on a \emph{written} phonetic
specification of the target word. For our rare words this specification
is absent, and where it exists their input reduces to the oracle-phoneme
condition (Table~\ref{tab:oracle}), which does not close the gap to
GRAFT's audio interface.

To isolate the contribution of the two key training-data choices, we
compare GRAFT against two ablations of itself. \emph{GRAFT (no~VC)}
slices the per-word hint clips directly from the source-speaker
recordings instead of voice-converted copies, with the cross-utterance
swap and augmentation disabled so no converted content re-enters.
\emph{GRAFT (no~aug.)} encodes the hints clean, without the silence
padding (Section~\ref{sec:data}) or the perturbations
$\mathrm{Aug}(\cdot)$ of Eq.~\eqref{eq:hint-sampling}.

\subsection{Metrics}
We use five objective metrics and one subjective metric. WER is the
sentence-level word error rate between the carrier text and the
Whisper-large-v3~\cite{whisper} transcript of the generated utterance,
capped at $1.0$ per item. D-PER, the main pronunciation metric, is the
phoneme error rate between the hint clip and the targeted word cut from
the generated audio: both are phonemised with the ZIPA universal phone
recogniser~\cite{zipa} and the Levenshtein distance between the phone
sequences is normalised by the reference length. Naturalness is
UTMOS~\cite{utmos}. Whisper similarity (WS) is the cosine similarity
between the Whisper-encoder embeddings of the targeted word in the
generated and hint clips, a content representation that largely discards
speaker identity, so WS reflects pronunciation rather than voice. Speaker preservation (SSIM) is
the WavLM~\cite{wavlm} cosine similarity to the cloned reference.
Finally, a blind pairwise listening study, run with an online
tool,\footnote{\url{https://www.mabyduck.com/}} gives per-system Bradley-Terry scores for
English (Fig.~\ref{fig:elo}).

\section{Results}
\label{sec:results}

\subsection{Difficult-word evaluation}
\label{sec:eval-difficult}

Table~\ref{tab:difficult} reports all metrics on the benchmark of
Section~\ref{sec:setup}, discussing English first and the other four
languages in Section~\ref{sec:eval-multilingual}. Because GRAFT shares
the base model's backbone and differs in the conditioning interface,
the GRAFT-versus-base comparison largely isolates the effect of that
interface.

\begin{figure}[!t]
\centering
\begin{tikzpicture}
\begin{axis}[
  ybar,
  width=\columnwidth,
  height=5.0cm,
  bar width=13pt,
  every axis plot/.append style={bar shift=0pt},
  ymin=1450, ymax=2330,
  xmin=0.3, xmax=9.7,
  ytick={1600,1800,2000,2200},
  xtick={1,2,3,4,5,6,7,8,9},
  xticklabels={GRAFT,XTTS-v2,F5-TTS,CosyVoice2,Qwen3-TTS,Matcha,Piper,MaskGCT,StyleTTS2},
  x tick label style={rotate=40,anchor=east,xshift=7pt,yshift=-5pt,font=\small},
  y tick label style={font=\footnotesize},
  ylabel={\small Bradley-Terry score ($95\%$ CI)},
  axis line style={draw=gray!55},
  tick style={draw=gray!55},
  tick pos=left,
  ymajorgrids, grid style={dashed,gray!22},
  error bars/error bar style={black!55,line width=0.8pt},
  clip=false,
]
\addplot+[draw=none,fill=graftclr,
          error bars/.cd,y dir=both,y explicit]
coordinates {
 (1,2197) += (0,71) -= (0,90)
};
\addplot+[draw=none,fill=baseclr,
          error bars/.cd,y dir=both,y explicit]
coordinates {
 (2,2047) += (0,75) -= (0,99)
 (3,2045) += (0,73) -= (0,95)
 (4,2030) += (0,80) -= (0,103)
 (5,1963) += (0,85) -= (0,113)
 (6,1958) += (0,79) -= (0,101)
 (7,1915) += (0,88) -= (0,110)
 (8,1830) += (0,98) -= (0,133)
 (9,1675) += (0,105) -= (0,128)
};
\end{axis}
\end{tikzpicture}
\caption{Blind pairwise human listening study on English (ratings initialised at
$1000$). Thirty difficult words from the benchmark, judged by $20$ paid listeners over
$600$ comparisons. With a clean recording of the target word as reference, each listener
was asked: \emph{``which of the two sentences has this pronunciation closer in terms of
phonetics, linguistics and prosody, while remaining natural?''} Bars are Bradley-Terry
scores, whiskers $95\%$ confidence intervals. GRAFT (green) ranks first, leading the next
system by $150$ points, with a lower bound ($2107$) above every baseline point estimate.}
\label{fig:elo}

\vspace{0.9em}
\small
\captionof{table}{Several per-word hints in one carrier, over $100$
multi-target carriers. $n$ is the grafted-word count, D-PER
averaged over them.}
\label{tab:multihint}
\renewcommand{\arraystretch}{0.98}
\begin{tabular*}{\columnwidth}{@{\extracolsep{\fill}}cccc@{}}
\toprule
\textbf{Hints}~$n$
  & \textbf{D-PER}\,$\downarrow$ & \textbf{WER}\,$\downarrow$ & \textbf{UTMOS}\,$\uparrow$ \\
\midrule
$1$ & $0.282$ & $0.049$ & $4.470$ \\
$2$ & $0.305$ & $0.117$ & $4.445$ \\
$3$ & $0.328$ & $0.222$ & $4.440$ \\
$4$ & $0.331$ & $0.245$ & $4.458$ \\
$5$ & $0.376$ & $0.206$ & $4.494$ \\
\bottomrule
\end{tabular*}
\end{figure}

On phonetic faithfulness GRAFT separates clearly from the baselines, and
its D-PER reduction over the base is significant in every language
(paired Wilcoxon over the $500$ words per language, $p<10^{-7}$). It also roughly halves the per-word D-PER
spread ($\pm0.16$ vs.\ $\pm0.42$ in English), so the correction is more
consistent, not only lower.
The Lingua Libre hints are reference single-word recordings by fluent
speakers, taken as ground-truth. Scoring against the hint invites two
objections. The acoustic one, that GRAFT pastes the hint through, is the
no-VC ablation: it reaches the \emph{lowest} D-PER ($0.16$,
Table~\ref{tab:ablation}) only by copying the hint while speaker
similarity collapses to $0.20$, a mode we train against. The metric one,
that D-PER phonemises both sides with one recogniser (ZIPA) and GRAFT
alone gets the hint as audio, so the score might reward channel match, is
ruled out by two judges sharing neither GRAFT's modality nor ZIPA: GRAFT
also leads on WS (a Whisper-encoder distance) and in the human study
(Fig.~\ref{fig:elo}, no phone metric). The benchmark thus measures
faithful transfer of the supplied pronunciation, the deployment case, not
canonical correctness. Giving the phoneme baselines the hint as phonemes
narrows but does not close the gap, and a human-curated dictionary
pronunciation does not either (Table~\ref{tab:oracle}, English
dictionary-covered words), ruling out transcription error as the cause
rather than settling the rarer-word case. Because WER is sentence-level, one mispronounced target word barely moves
it, so a system can post a low WER while still getting the word wrong.
GRAFT's higher WER is itself concentrated on the target word, removing it
drops the English WER from $0.09$ to $0.03$ (base $0.02$ to $0.01$) and
the five-language average from $0.09$ to $0.05$, leaving the carrier over
$95\%$ correct. The residual is partly the recogniser's, even on a clean
isolated human recording, transcribed in isolation, Whisper-large-v3~\cite{whisper} returns the
exact word only $75\%$ of the time in English ($66\%$ overall), and on
over half of GRAFT's misses the rendered pronunciation still matches the
hint (D-PER${\le}0.34$). Naturalness is largely unaffected:
GRAFT leads the base on UTMOS in four of five languages.

The ablations (Table~\ref{tab:ablation}) isolate voice-conversion
training. To test whether the grafted word merely echoes the hint
speaker, the generated target word is cut from each render with the same
forced aligner, resampled to $16$~kHz, and scored with
ECAPA-TDNN~\cite{ecapa} against both the hint speaker and the cloned
target speaker, averaged over the benchmark items. For the no-VC ablation
this confirms the paste-through as the grafted word sits on the hint speaker
($0.67$) far more than on the target ($0.10$). With voice conversion the
grafted word instead carries the target identity rather than the hint's
($0.47$ versus $0.12$), which a copy of the hint waveform cannot produce
and which separates GRAFT's low D-PER from acoustic leakage. 

\subsection{Multilingual evaluation}
\label{sec:eval-multilingual}
The hint conditions on acoustic tokens rather than text or phonemes, so
the interface carries no language-specific machinery. The non-English
rows of Table~\ref{tab:difficult} repeat the English pattern as GRAFT cuts target-word D-PER by $22$--$39\%$ in every language and raises hint similarity, with naturalness and speaker similarity preserved at a small WER cost.

\begin{table}[t]
\centering
\caption{Oracle-pronunciation control (English CMUdict-covered words,
$n=225$). Phoneme baselines receive either the wav2vec2-espeak
transcription of the hint or the human-curated CMUdict pronunciation.
Neither closes the gap to GRAFT, which rules out transcription error as the
cause of the baseline deficit on these covered words. D-PER is scored
against the hint, which the audio-derived phonemes track directly.}
\label{tab:oracle}
\footnotesize
\setlength{\tabcolsep}{8pt}
\renewcommand{\arraystretch}{1.05}
\begin{tabular}{@{}lcc@{}}
\toprule
\textbf{System} & \textbf{hint phonemes}\,$\downarrow$ & \textbf{oracle (CMUdict)}\,$\downarrow$ \\
\midrule
StyleTTS2 & $0.65$ & $0.69$ \\
MaskGCT   & $0.49$ & $0.45$ \\
Matcha-TTS & $0.43$ & $0.49$ \\
Piper     & $0.49$ & $0.51$ \\
\midrule
\textbf{GRAFT} (audio) & \multicolumn{2}{c}{$\mathbf{0.30}$} \\
\bottomrule
\end{tabular}
\end{table}

\section{Discussion}
\label{sec:discussion}

The human study is GRAFT's strongest evidence, a human judgment rather than
a phone-distance score. The reference shown to raters is
the same recording GRAFT receives as its hint, so the study measures which model reproduces a supplied pronunciation most faithfully, not which is canonically correct. Every system is given that same pronunciation, as audio for GRAFT and as transcribed phonemes or spelling for the others. Across $30$ difficult words and $600$ comparisons GRAFT wins outright, leading the strongest baseline by $150$ Bradley-Terry points with its confidence-interval lower bound above every other system's point estimate. It does so not by echoing the clip but by re-rendering the pronunciation in the cloned target voice (the per-word identity above), so the preference reflects pronunciation transfer, not playback of the hint.

Because the hint is an example rather than a transcription, a non-expert
can correct a chosen word by saying it, and disentanglement keeps the output
in the chosen voice. This helps where text and phoneme front-ends fail:
rare proper nouns, loanwords and technical terms whose realisation is
known to a speaker yet absent from any lexicon.

Training grafts every eligible word (Section~\ref{sec:data}), so several
hints stack in one carrier: up to five (Table~\ref{tab:multihint}) raise
per-word D-PER only gently ($0.28$ to $0.38$) while carrier WER absorbs
the seam cost ($0.05$ to over $0.20$) and naturalness is unchanged, so
the price of extra grafts falls on fluency, not pronunciation.

The method is reliable when the hint matches the trimmed, neutral
training distribution, and mistrimmed or low-quality hints can yield a
wrong pronunciation. The interface operates on
single words, and GRAFT is shown on one $0.6$B base model. The benchmark
is non-tonal, so the measured gains reflect segmental and stress detail
rather than the tonal advantage motivated against symbolic phonemizers,
which discard diacritics. Multi-word spans, prosody and tonal-language
transfer, where such phonemizers should fail most, and scaling are left
to future work.

\section{Conclusion}
GRAFT adds a per-word audio conditioning slot to neural codec language
model TTS: a short spoken example fixes a word's pronunciation, and
voice-conversion training renders it in the cloned voice from any speaker.
The mechanism reuses the base model's own tokenizer and adds no new
parameters, so per-word control drops into an existing system at the cost
of only a few extra tokens at inference. Because the interface is a spoken
example rather than a symbolic transcription, a non-expert can prescribe a
pronunciation simply by saying the word, which is precisely what rare
proper nouns, loanwords and technical terms demand and what text and
phoneme front-ends cannot supply.

Human raters rank GRAFT first in a blind English listening study, and on
the objective benchmark it improves target-word pronunciation fidelity
over the text-only base (D-PER reduced by $22$--$39\%$) with consistent
gains across five languages, while preserving speaker similarity and
naturalness. We release the checkpoints, code and the multilingual
difficult-word benchmark to support further work. Tonal-language transfer,
multi-word spans, and scaling to larger backbones, where symbolic
front-ends should struggle most, are natural next steps.

\bibliographystyle{IEEEtran}
\bibliography{refs}

\end{document}